\pdfoutput=1

\documentclass[11pt]{article}

\usepackage[preprint]{acl}

\usepackage{times}
\usepackage{latexsym}
\usepackage{subfiles}
\usepackage{xurl}

\usepackage{tabularx}
\usepackage{algorithm2e}

\usepackage{amsmath}

\usepackage[T1]{fontenc}

\usepackage[utf8]{inputenc}

\usepackage{microtype}

\usepackage{inconsolata}

\usepackage{array}
\usepackage{makecell} 
\usepackage{multirow}
\usepackage{graphicx}
\usepackage{cleveref}
\usepackage{acronym}
\usepackage{subcaption}
\acrodef{nlp}[NLP]{Natural Language Processing}
\acrodef{lm}[LM]{Language Model}
\acrodef{wsd}[WSD]{Word Sense Disambiguation}
\acrodef{lane}[LANE]{Lexical Adversarial Negative Example}
\acrodef{lscd}[LSCD]{Lexical Semantic Change Detection}

\title{LANE: Lexical Adversarial Negative Examples\\ for Word Sense Disambiguation}

\author{
 \textbf{Jader Martins Camboim de S\'a\textsuperscript{1,2}},
 \textbf{Jooyoung Lee\textsuperscript{1,3}},
 \textbf{C\'edric Pruski\textsuperscript{2}},
 \textbf{Marcos Da Silveira\textsuperscript{2}}\\
 \textsuperscript{1}FSTM - University of Luxembourg\\
  2 place de l’Université, L-4365, Esch-sur-Alzette, Luxembourg,\\
 \textsuperscript{2}Luxembourg Institute of Science and Technology\\
        5 avenue des Hauts-Fourneaux, L-4362, Esch-sur-Alzette, Luxembourg,\\
\textsuperscript{3}Brown University\\
  Providence, RI 02912, United States,\\
 \small{
   \textbf{Correspondence:} \href{mailto:email@domain}{first.second@list.lu}
 }
}

\begin{document}
\maketitle
\begin{abstract}
Fine-grained word meaning resolution remains a critical challenge for neural language models (NLMs) as they often overfit to global sentence representations, failing to capture local semantic details. We propose a novel adversarial training strategy, called LANE, to address this limitation by deliberately shifting the model's learning focus to the target word. This method generates challenging negative training examples through the selective marking of alternate words in the training set. The goal is to force the model to create a greater separability between same sentences with different marked words. Experimental results on lexical semantic change detection and word sense disambiguation benchmarks demonstrate that our approach yields more discriminative word representations, improving performance over standard contrastive learning baselines. We further provide qualitative analyses showing that the proposed negatives lead to representations that better capture subtle meaning differences even in challenging environments. Our method is model-agnostic and can be integrated into existing representation learning frameworks.

\end{abstract}

\section{Introduction}
\label{sec:intro}
\ac{wsd}, the task of identifying the precise meaning of a word in context, remains central to deep semantic understanding. It underpins applications such as \ac{lscd} \cite{Sa2024SurveyIC} and word similarity tasks \cite{armendariz-etal-2020-semeval}. For example, \ac{wsd} systems aim to determine whether the word ``crazy'' denotes insanity (as in \textit{That’s a crazy man}) or excitement (as in \textit{That's crazy, man}), and whether ``bank'' refers to a financial institution or a river edge. 
While these words can appear in similar contexts, its particular usage can modulate completely different meanings.

Modern neural language models like XLM-Roberta achieve impressive results in this task, but their final word representation often overfit to global context rather than encoding a word's specific sense \cite{liu-etal-2021-am2ico}. They solve the task by capturing topical cues, for instance, inferring that ``bank,'' ``loan,'' and ``interest'' signal finance, but can misrepresent a word’s fine-grained contribution \cite{Xu_2025} \cite{mccoy-etal-2019-right}. This limitation surfaces in cases like heavy rain vs. heavy traffic: both imply ``a lot,'' yet with different nuances, intensity/volume versus density/severity. Such failures to capture precise contextual meaning hinder performance in downstream tasks requiring genuine semantic nuance, like neologism identification \cite{mccrae-2019-identification}.

\begin{figure}[htpb]
    \centering
    \includegraphics[width=\linewidth]{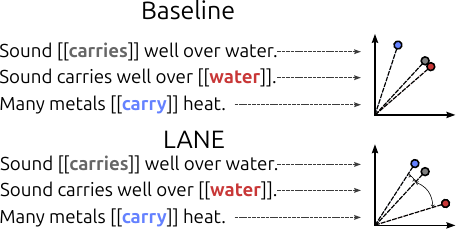}
    \caption{Comparison of traditional (Baseline) and \ac{lane} learned representations. Colors represent anchor (gray), positive (blue), and adversarial negative (red).}
    \label{fig:lane}
\end{figure}

A widely adopted strategy to improve lexical sensitivity in \ac{wsd} is target-word highlighting, where the word of interest is marked (e.g., with special tokens) before being encoded by a language model \cite{cassotti-etal-2023-xl}. The underlying assumption is that explicit marking encourages the model to attend more directly to the lexical semantics of the word. Yet, as our analysis reveals (\Cref{fig:lane}), this assumption is fragile: embeddings of the same sentence remain nearly identical regardless of which word is highlighted, indicating that the model often resolves the task using only the surrounding context \cite{liu-etal-2021-am2ico}.

In practice, this results in nearly indistinguishable representations for semantically distinct words, while instances of the same sense across different contexts may be mapped to more distant vectors. Such behavior illustrates the persistence of shortcut learning \cite{Robinson2021CanCL}, where models exploit superficial contextual regularities rather than grounding their predictions in the lexical meaning of the marked token.

This imbalance produces a collapsed representation space, where embeddings concentrate in a narrow region irrespective of the highlighted word. The consequence is reduced lexical separability and weak generalization on tasks requiring sensitivity to word-level sense, especially in contexts that are semantically similar. Taking this into consideration, in this paper, we investigate \textit{How new approaches can ensure that the representation space meaningfully reflects the contribution of the target word?}

To address this, we introduce the LANE framework, which biases representations toward lexical relevance. LANE generates adversarial negatives by substituting the highlighted word while keeping the sentence context unchanged. These hard negatives share the full contextual frame of the anchor sentence but differ in the target token, forcing the model to focus on the word’s semantics rather than relying solely on surrounding context.

By presenting nearly identical sentences that differ only in the marked word, LANE sharpens semantic boundaries and produces more discriminative representations. This encourages the model to attend precisely to the lexical contribution of the target word, resulting in embeddings that better capture fine-grained word meaning.

We evaluate \ac{lane} across both English and multilingual datasets, spanning a diverse set of architectures. Results show consistent improvements in lexical representation quality and cross-domain generalization, demonstrating that targeted lexical supervision can enhance robustness even in high-resource, context-rich settings. Furthermore, the method’s agnostic nature allows it to be incorporated into existing training pipelines with minimal computational overhead or implementation complexity, making it broadly applicable for multilingual lexical tasks.

\section{Related Work}
\label{sec:related}
Representing fine-grained word meaning has long been a longlasting challenge in natural language processing. Early approaches tackled this problem through lexical resources such as WordNet \cite{Miller1995WordNetAL} and the creation of sense-annotated corpora to support supervised \ac{wsd} methods \cite{raganato-etal-2020-xl} \cite{Bevilacqua2021RecentTI} \cite{huang-etal-2024-sense}. With the advent of deep learning, contextualized word embeddings such as ELMo \cite{peters-etal-2018-deep} and BERT \cite{devlin-etal-2019-bert} became standard, enabling substantial gains in \ac{wsd} and related tasks by dynamically adjusting representations according to context.

Yet despite these advances, contextualized models often fail to distinguish subtle word-level meaning differences, tending instead to overfit to sentence-level semantics \cite{ethayarajh-2019-contextual}. This weakness is particularly visible in \ac{lscd}, which requires tracking the semantic shifts of individual words across time and domains. While static embedding approaches offered initial baselines \cite{schlechtweg-etal-2019-wind} \cite{martinc-etal-2020-discovery}, more recent work has leveraged contextualized embeddings to capture dynamic variation \cite{schlechtweg-etal-2020-semeval} \cite{kutuzov-giulianelli-2020-uio} \cite{giulianelli-etal-2020-analysing}. Nevertheless, isolating the semantics of the target word from the broader discourse remains an open challenge.

To address these limitations, recent research has turned to contrastive learning as a way to induce more discriminative semantic representations. Inspired by advances in computer vision \cite{Chen2020ASF}, contrastive objectives have been adapted to NLP, yielding stronger sentence and word embeddings for tasks such as semantic similarity and clustering \cite{gao-etal-2021-simcse}.

Within lexical semantics, several directions have emerged. One line of work leverages multilingual pretraining on large-scale, diverse corpora to improve contextual coverage of word senses \cite{cassotti-etal-2023-xl} \cite{Yadav2025XLDURelFS}. Another incorporates auxiliary supervision, such as in-context sense induction, to encourage sense separation \cite{giulianelli-etal-2023-interpretable} \cite{mosolova-etal-2024-injecting} \cite{lietard-etal-2024-word} \cite{Lietard2025CALEC}. A third explores data augmentation, for example by altering input structure: Martelli et al. proposed swapping sentence order in a cross-encoder \cite{martelli-etal-2021-semeval}, though this strategy does not extend to bi-encoder architectures where contrastive losses are typically computed.

Complementary to these efforts, adversarial strategies aim to construct more challenging negatives to prevent representational collapse. Prior work has explored lightweight perturbations such as case alternation \cite{Wang2022ImprovingCL}, synonym and antonym substitution \cite{wang-etal-2021-cline}, and token replacements with masked language model predictions \cite{chuang-etal-2022-diffcse}. Multilingual adversarial signals have further been obtained through cross-lingual links in Wikipedia, which align English sentences with low-resource counterparts while introducing cross-lingual semantic contrasts \cite{liu-etal-2021-am2ico}. While these methods increase robustness and sharpen decision boundaries, they typically operate at the sentence level, focusing on global semantic differences rather than directly enforcing separability of meanings for a specific word in context.

Our work addresses this gap by combining the strengths of contrastive learning and adversarial augmentation while targeting their shared limitation: the absence of adversarial signals that operate at the word level. Instead of relying on heuristic perturbations or external lexical resources, we generate adversarial negatives by marking different words within the same sentence, producing pairs that are contextually identical yet lexically distinct. This strategy is model-agnostic and integrates seamlessly into existing frameworks for representation learning.

In doing so, it complements multilingual pretraining and auxiliary-task-based sense induction, while providing adversarial challenges directly tied to the phenomenon of interest: the fine-grained disambiguation of word meaning. Our experiments on both \ac{wsd} and \ac{lscd} detection confirm that this approach yields more discriminative word representations than standard contrastive and adversarial baselines, underscoring the importance of aligning adversarial objectives with lexical semantics.

\section{Datasets}
\label{sec:datasets}
Evaluating language models on a single dataset has long been standard practice in \ac{nlp}, yet this evaluations often provide an incomplete view of generalization and robustness \cite{Lones_2024}. Even within a single task, datasets may differ substantially in domain, linguistic complexity, annotation conventions, and underlying assumptions. Consenquently, relying on one benchmark risks overfitting to dataset-specific idiosyncrasies rather than assessing a model's true capacity for semantic discrimination in \ac{wsd}.

To obtain a more comprehensive evaluation, we curated four datasets covering diverse sources and temporal spans: SEMCOR \cite{miller-etal-1993-semantic}, MASC\footnote{\url{https://anc.org/data/masc/}}, FEWS \cite{blevins-etal-2021-fews}, and DWUG \cite{schlechtweg-etal-2024-dwugs}. These datasets differ primarily along two dimensions: (i) time of creation (i.e. 1993 vs. 2020) and (ii) source type (i.e. dictionaries, books, blogs). This diversity enables controlled variation in language register and data quality (ranging from the standardized style of dictionaries to the more informal, conversational tone of blogs) thus exposing models to a broad spectrum of lexical and contextual phenomena.

\begin{table*}[htbp]
    \centering
    \caption{Examples of sentences pairs in the WiC dataset and corresponding hard\_negative examples generated by LANE}
    \label{tab:tab_datasets}
    \begin{tabularx}{\textwidth}{c c X X}
    \hline
    \textbf{Dataset} & \textbf{Label} & \textbf{Sentence 1} & \textbf{Sentence 2} \\
    \hline
    SEMCOR (1993) & positive & The heavens opened, pelting them with \emph{hail} the size of walnuts. & Drought, \emph{hail}, disease, and insects take their toll of crops. \\
    MASC (2003)   & negative & The report wasn't hard, I had already read the book, and so I just jotted down a few \emph{notes} for him. & Passing \emph{notes} when the teacher isn't paying attention. \\
    FEWS (2021)   & positive & Virchow stated that premature fusion of this suture results in \emph{pachycephalic} deformity. & [...] that a judge may be learned in the law, but woefully \emph{pachycephalic} in matters scientific. \\
    \hline
    \end{tabularx}
\end{table*}

Additional to the created datasets we evaluate our models in WiC data \cite{pilehvar-camacho-collados-2019-wic} and DWUG \cite{schlechtweg-etal-2021-dwug}. While WiC and DWUG are formatted for \ac{wsd} formulations, SEMCOR, MASC, and FEWS required preprocessing as they just present usages and sense keys. We generated contrastive pairs by merging instances with identical lemma and POS tags, then pairing them across distinct context. This process yielded pairs of the same lexical item in different environments, supporting fine-grained semantic discrimination. Ground-truth labels were automatically assigned using sense keys: pairs sharing the same key were labeled as positive (1), while those differing in sense were labeled as negative (0). This alignment to sense inventories ensures that the resulting pairs genuine semantic distinctions rather than superficial contextual variation.

For SEMCOR, MASC, and FEWS, we partitioned data lexicographically such that test sets contain words beginning with letters ``P'' or later. This reduces lexical overlap between training and testing splits, forcing models to generalize to unseen lexical items. Unlike DWUG and WiC, our resources include adjectives and adverbs in addition to nouns and verbs, thereby broadening the coverage of syntactic categories and semantic phenomena.

For multilingual evaluation, we adopt XL-WiC \cite{raganato-etal-2020-xl} which extends WiC's word in context formulation to multiple languages (see \Cref{tab:tab_multilingual}). XL-WiC draws on both WordNet and Wiktionary to maintain cross-lingual consistency. Additionally, we train with XL-LEXEME, a composite resource integrating XL-WiC, MCL-WiC, and AM2iCO, to further enhance cross-lingual and cross-categorical representativeness.

The Word-in-Context task requires models to distinguish whether two target word usages convey the same meaning. A common learning paradigm for this task is contrastive learning \cite{cassotti-etal-2023-xl}, where a sentence ($x$) serves as an anchor, paired with positive $(x,x^+)$ or negative $(x,x^-)$ examples. In \ac{wsd}, the anchor contains a target word appearing in two contexts: in positive pairs, the meanings coincide; in negative pairs, they differ. When encoding these pairs with Transformer models, target words are marked explicitly, either via prefix notation (e.g., \texttt{word<s>context}) or inline markers (e.g., \texttt{<t>word</t>} in the sentence). However, such marking schemes can cause the model to rely on sentence-level cues rather than the target word itself \cite{liu-etal-2021-am2ico}. To address this, our mehtod (LANE) generates hard-negative examples by varying the marked token within identical sentence contexts, thereby enforcing a stronger focus on the target word (see \Cref{tab:tab_example}).

\begin{table}[htbp]
    \centering
    \caption{Examples of positive, negative, and hard negative pairs in WiC. Hard negatives are generated by LANE.}
    \label{tab:tab_example}
    \begin{tabularx}{\linewidth}{c X X}
    \hline
    \textbf{Label} & \textbf{Sentence 1} & \textbf{Sentence 2} \\
    \hline
    positive      & Sound \emph{carries} well over water. & Many metals \emph{carry} heat. \\
    negative      & Sound \emph{carries} well over water. & You must \emph{carry} your camping gear. \\
    hard negative & Sound \emph{carries} well over water. & Sound carries well over \emph{water}. \\
    \hline
    \end{tabularx}
\end{table}

\section{Methodology}
\label{sec:method}
Ideally, an embedding space for word senses should map all occurrences of the same sense to a single, consistent vector regardless of context. In practice, this entails positioning anchor–positive pairs (same sense, different contexts) close together in the embedding space, while pushing anchor–negative pairs (different senses) apart, a principle often formalized through the ball-packing problem \cite{Robinson2020ContrastiveLW}.

We propose a revised objective that goes beyond distinguishing positive and negative pairs by enforcing self-differentiation. In this setting, the surrounding sentence remains identical, but the marked target word differs. To operationalize this, we introduce adversarial negative examples generated through a rule-based procedure that randomly replaces the highlighted word in the sentence with another lexical candidate drawn from the training data (\Cref{algo:lane}). These adversarial negatives prevent the model from relying solely on contextual cues and instead compel it to attend to the lexical identity of the target word. During training, adversarial negatives are integrated alongside standard positive and negative pairs, enhancing the model’s ability to learn fine-grained word sense representations.

In the following section, we describe how these adversarial negatives are generated dynamically and progressively replace a small proportion of the training data over epochs.

\subsection{Adversarial Negative Examples}
Bi-encoders independently represent vectors in the embedding space without explicit awareness of which target word the model should attend to. Although marking tokens are intended to signal the target, they have minimal effect in practice (\Cref{sec:eval}). To improve word-sense learning, we aim to reduce the model’s reliance on contextual cues during training. To this end, we generate adversarial negative examples with two desirable properties:

\begin{itemize}
    \item \textbf{Property 1:} A contextualized word representation should produce a distinct vector whenever the marked lexical item changes, even if the surrounding context remains identical.

    \item \textbf{Property 2:} For words $X$ in sentence $A$ and $Y$ in sentence $B$ that share the same sense, their embeddings should remain similar only when $X$ and $Y$ are the marked items; the embeddings should diverge whenever a different lexical item is marked in $A$ or $B$.
\end{itemize}
The vector representation of a word should change whenever a different lexical item in the same sentence is marked as the target (Property 1), following the assumption of avoidance of repetition \cite{walter2007repetition}. When two words share the same sense across different sentences, their embeddings should diverge if another word is marked (Property 2).

Building on these properties, we generate adversarial examples that explicitly encode this learning constraint. During training of each dataset, we construct: (i) pairs of identical sentences that differ only in which word is marked as the target; and (ii) pairs in which sentences $A$ and $B$ contain the same sense for a given word, but a different word in $A$ is marked. This construction ensures that even identical or highly similar contexts yield distinct representations when the lexical target changes.

In \Cref{algo:lane}, we illustrate our method for dynamically generating negative examples in-batch. The input consists of a dictionary containing two target words (word1, word2), two corresponding sentences (sentence1, sentence2), and a similarity label (label):

\begin{itemize}
    \item First, the function extracts all tokenized words from `sentence1' and filters out any occurrences of `word1'.
    \item From the remaining tokens, one word is randomly selected as newword.
    \item If the original `label = 0.0' (indicating dissimilarity between the pair), the output replaces `word2' with `newword' while making both sentences identical to `sentence1'.
    \item Otherwise (`label $\neq$ 0.0'), the function replaces `word1' with `newword' while keeping `sentence2' unchanged.
\end{itemize}

In both cases, the newly generated example is assigned a label of `0.0', ensuring that it serves as a challenging negative instance. Because the contexts remain nearly identical, the model is forced to focus on the marked lexical item rather than relying on the surrounding context. We note that the likelihood of `newword' sharing the original meaning is low, consistent with the \textit{avoidance of repetition} assumption.

\begin{algorithm}[htbp]
\caption{Pseudo-code for creating adversarial negative examples (\texttt{lexical-negative})}
\label{algo:lane}
\KwIn{$w_1, s_1, w_2, s_2, label$}
\KwOut{$w_1, s_1, w_2, s_2, label$}

$words \gets \text{split-into-words}(s_1)$\;
$candidates \gets \{\, w \mid w \in words,\, w \neq w_1 \,\}$\;
$newword \gets \text{random-choice}(candidates)$\;

\If{$label = 0$}{
    $w_2 \gets newword$\;
    $s_2 \gets s_1$\;
}
\Else{
    $w_1 \gets newword$\;
    $label \gets 0$\;
}

\Return{$w_1, s_1, w_2, s_2, label$}\;
\end{algorithm}

We apply a multilingual word splitter to accommodate languages such as Farsi and Japanese, where whitespace is not a reliable delimiter. Finally, to prevent the model from overfitting to these challenging negatives or collapsing into a local optimum, we introduce the adversarial examples gradually through a scheduled insertion strategy.

\subsection{Scheduler}
Introducing adversarial negatives too early in training risks convergence to suboptimal local minima (\Cref{sec:schedule}). In such cases, the model may overfit to a narrow subset of challenging comparisons before developing robust and generalizable representations \citep{Xuan2020HardNE}.

To mitigate this issue, we employ a linear scheduling strategy, scaled by the training epoch, to gradually introduce adversarial negatives. During the initial warm-up phase, the contrastive language model is trained without any adversarial examples, allowing it to form stable base representations. After this stage, the probability of sampling adversarial examples increases linearly with each epoch, ensuring a smooth and controlled transition from easy to hard comparisons. This progressive introduction balances representational stability with increasing task difficulty. All negatives are computed dynamically from in-batch data rather than precomputed, enabling the model to adapt continuously as training evolves.

Our ablation studies corroborate the effectiveness of this strategy (\Cref{sec:schedule}): the linear scheduler prevents early collapse, enhances training stability, and achieves superior overall performance compared to settings where adversarial examples are introduced prematurely or at a fixed rate.

\subsection{Optimization Objective}
To differentiate senses, we optimize our models with a contrastive learning objective. Following recent literature in lexical differentiation, we employ an in-batch, cosine–based loss \cite{Yadav2025XLDURelFS}. This formulation leverages implicit negatives within each batch, thereby promoting semantically coherent clustering via cosine similarity. Our choice aligns with prior state-of-the-art methods and provides a strong, empirically validated baseline. All models are optimized using the CoSENT loss \cite{Huang2024CoSENTCS}, as formalized in the equation below:

\begin{equation}
    \mathcal{L} = \log \left(1 + \sum_{i,j\hspace{0.5em} y_i < y_j} e^{ \lambda (s_i - s_j)} \right)
\end{equation}

Here, $s_k$ denotes the cosine similarity score for the $k$-th embedding pairs such that the expected similarity of $s_i$ is greater than $s_j$. The summation extends over all ordered pairs in the batch such that $y_i < y_j$. If the embeddings are $u_k$ and $v_k$, then $s_k = \cos(u_k, v_k) = \frac{u_k \dot v_k}{|u_k| |v_k|}$. $y_k$ represents the ground-truth similarity label for the $k$-th pair, and $\lambda$ is a trainable scaling factor that controls the sharpness of the distribution.

If the model ranks correctly ($s_j > s_i$), the difference $(s_i - s_j)$ is negative, and the exponential term $e^{\lambda (s_i - s_j)}$ becomes small, contributing minimally to the overall loss. Conversely, when the model ranks incorrectly ($s_i > s_j$), the difference $(s_i - s_j)$ is positive, causing $e^{\lambda (s_i - s_j)}$ to grow large and yield a higher loss value. This formulation thus penalizes misranked pairs more heavily while suppressing contributions from correctly ranked ones. In all experiments, we use the standard scaling parameter $\lambda = 20$.

\section{Evaluation}
\label{sec:eval}
We assess the effectiveness of the proposed LANE method on the datasets described in \Cref{sec:datasets}. Each dataset is split into train, development, and test partitions. During training, negative mining is applied exclusively to encourage more discriminative word representation, while the development split is used for model selection, and the test split is reserved for final evaluation. 

To quantify the added value of LANE, we conduct controlled comparisons using two modern transformer architectures as backbone models: DeBERTa-v3 (DV3) and ModernBERT (MBERT), and the established architecture XLM-RoBERTa (XLM-R). For each architecture, we evaluate both the baseline model trained with standard contrastive objectives and the same model augmented with LANE. This design isolates the impact of LANE on learning fine-grained word-sense distinctions, independent of the underlying architecture.

All models are trained using AdamW optimizer \cite{Loshchilov2017FixingWD} with a learning rate of 1e-5, 500 warm-up steps, a weight decay of 0.01, and an effective batch size of 64. Models with LANE are trained for 20 epochs, while baseline models without LANE are trained for 10 epochs, given its early convergence, with model selection based on performance on the development split. This evaluation framework ensures a fair and direct comparison, allowing us to highlight improvements in word-sense representation and downstream performance attributable specifically to LANE.

\begin{table}[htbp]
\centering
\caption{Comparison of methods across datasets in terms of accuracy for test data.}
\resizebox{\linewidth}{!}{%
\begin{tabular}{l c c c c c}
\hline
\textbf{Model} & \textbf{WiC} & \textbf{DWUG} & \textbf{SEMCOR} & \textbf{MASC} & \textbf{FEWS} \\
\hline
MBERT & 0.567 & \textbf{0.735} & 0.722 & 0.733 & 0.509 \\
MBERT+\textbf{LANE} & \textbf{0.589} & 0.734 & \textbf{0.756} & \textbf{0.759} & \textbf{0.523} \\
DV3 & 0.655 & 0.735 & \textbf{0.759} & 0.754 & 0.629 \\
DV3+\textbf{LANE} & \textbf{0.660} & \textbf{0.737} & 0.756 & 0.754 & 0.629 \\
XLM-R & 0.705 & 0.739 & \textbf{0.759} & 0.752 & 0.627 \\
XLM-R+\textbf{LANE} & \textbf{0.721} & \textbf{0.742} & \textbf{0.759} & \textbf{0.773} & \textbf{0.647} \\
\hline
\end{tabular}}
\label{tab:tab_english}
\end{table}

As shown in \Cref{tab:tab_english}, incorporating \ac{lane} adversarial negatives consistently improves, or at minimum maintains, accuracy across all evaluated datasets. For instance, MBERT sees notable gains on WiC, SEMCOR, MASC and FEWS when combined with LANE, highlighting its effectiveness in enhancing word-sense discriminability. Similarly, DeBERTa-v3 exhibits improvements with LANE, particularly on the WiC and DWUG datasets, demonstrating that even strong monolingual baselines benefit from the adversarial word-focused training signal. 

Interestingly, XLM-RoBERTa, despite being older than DeBERTa-v3 and MBERT, achieves the highest baseline accuracy across several datasets, reflecting its strong pretraining on large-scale multilingual data. When augmented with LANE, XLM-R not only improves further, most noticeably on WiC, MASC, and FEWS, but also attains the overall highest scores across the board, confirming the robustness of LANE across architectures.

These results indicate that LANE consistently enhances word-level representations, improving performance on diverse datasets regardless of the underlying model architecture. While DeBERTa-v3 achieves performance comparable to XLM-R on monolingual datasets, its monolingual design restricts its applicability in multilingual settings, as further illustrated in Table \ref{tab:tab_multilingual}. Overall, the evaluation demonstrates that LANE provides a substantial and architecture-agnostic boost to contrastive learning of fine-grained word senses.

Building on the monolingual experiments, we further evaluate our approach in a multilingual word sense disambiguation setting using the XL-LEXEME datasets (XL-WiC \cite{raganato-etal-2020-xl}, MCL-WiC \cite{martelli-etal-2021-semeval}, AM2ICO \cite{liu-etal-2021-am2ico}). 

We denote our model as XL-WiC+LANE (Discriminative Enhanced Lexical Training with Adversarial negatives) and have released it on HuggingFace\footnote{Omitted for review.}. \Cref{tab:tab_multilingual} reports our results on the XL-WiC test set, comparing XL-WiC+LANE with the base XL-LEXEME model and prior state-of-the-art methods.
As shown in the table, incorporating LANE consistently improves performance across most languages. XL-WiC+LANE achieves notable gains in English, French, German, Bulgarian, Chinese, and Dutch, demonstrating that adversarial word-level negatives enhance cross-lingual word-sense discrimination. These results confirm that LANE provides a measurable benefit even when applied to strong multilingual baselines.
We further assess our approach in an unsupervised cross-lingual generalization scenario using XL-WiC across nine target languages: Bulgarian (bg), Chinese (zh), Croatian (hr), Danish (da), Dutch (nl), Estonian (et), Farsi (fa), Japanese (ja), and Korean (ko). The model is trained solely on XL-LEXEME data and compared to XL-DUREL, which leverages substantially more training resources. Despite this difference, XL-WiC+LANE achieves comparable performance, highlighting its data efficiency and robust cross-lingual generalization.
Overall, these experiments extend the findings from the monolingual evaluation: LANE consistently enhances word-level representations, providing performance gains across languages, and supporting effective sense disambiguation in both supervised and cross-lingual settings.

\begin{table*}[htbp]
\centering
\caption{Comparison of classifiers with LANE for supervised multi-lingual (XL-WiC).}
\label{tab:tab_multilingual}
\resizebox{\textwidth}{!}{%
\begin{tabular}{l c c c c c c c c c c c c c c}
\hline
\textbf{Model} & \textbf{en} & \textbf{fr} & \textbf{de} & \textbf{it} & \textbf{bg} & \textbf{zh} & \textbf{hr} & \textbf{da} & \textbf{nl} & \textbf{et} & \textbf{fa} & \textbf{ja} & \textbf{ko} & \textbf{MEAN} \\
\hline
XL-LEXEME & \underline{0.722} & \underline{0.785} & 0.848 & \textbf{0.756} & \underline{0.827} & \underline{0.802} & 0.727 & \textbf{0.766} & 0.782 & 0.664 & \underline{0.666} & 0.682 & 0.798 & \underline{0.755} \\
XL-DUREL\textsuperscript{**} & 0.732 & 0.778 & \underline{0.850} & 0.729 & 0.754 & 0.777 & \textbf{0.752} & 0.756 & \underline{0.792} & \underline{0.676} & \textbf{0.706} & \textbf{0.697} & \textbf{0.801} & 0.753 \\
XL-WiC+\textbf{LANE} & \textbf{0.734} & \textbf{0.804} & \textbf{0.871} & \underline{0.746} & \textbf{0.847} & \textbf{0.805} & \underline{0.735} & \textbf{0.766} & \textbf{0.800} & \textbf{0.684} & 0.642 & \underline{0.683} & \underline{0.799} & \textbf{0.762} \\
\hline
\end{tabular}%
}
\end{table*}

\section{Discussion}
\label{sec:discussion}
Our results highlight a persistent limitation of contextual language models: even highly capable encoders often rely on global sentence semantics rather than representing the fine-grained meaning of the target word. Consistent with prior work \cite{liu-etal-2021-am2ico}, we observe that embeddings of the same sentence remain nearly identical regardless of which word is marked, indicating that lexical information contributes little to the learned representation. This reliance on coarse contextual cues undermines interpretability and restricts generalization to new domains or languages where such cues differ.

\begin{figure}
    \centering
    \includegraphics[width=\linewidth]{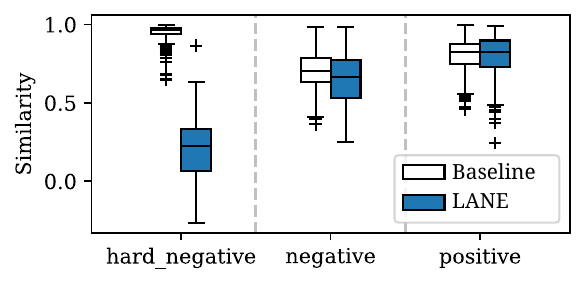}
    \caption{Similarity for representations learned under a traditional regime and under \ac{lane}.}
    \label{fig:boxplot}
\end{figure}

The proposed LANE framework mitigates this issue by introducing lexically controlled adversarial negatives—sentences that differ only in the marked token. This formulation constrains the learning process to capture distinctions that are attributable to the target word itself. As shown in Figure \ref{fig:boxplot}, this results in more structured embedding spaces: sentences containing the same sense are drawn closer together, while those differing in meaning are pushed farther apart. The improved intra-sense cohesion and inter-sense separation explain the consistent accuracy gains reported in the evaluation.

\begin{figure}
    \centering
    \includegraphics[width=\linewidth]{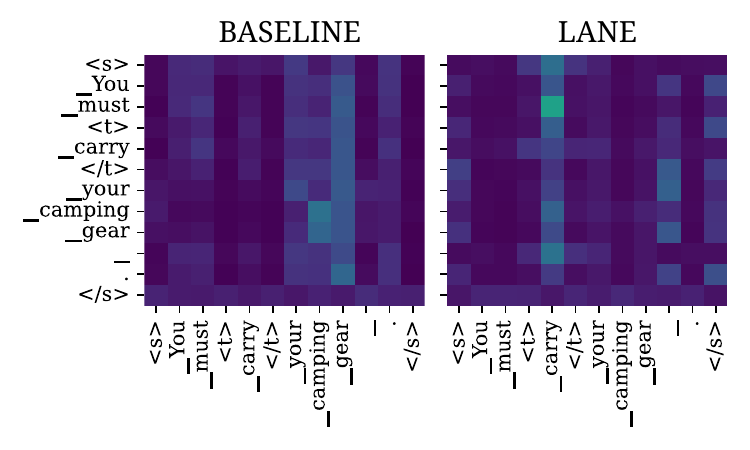}
    \caption{Last attention head heatmap for the baseline and \ac{lane}.}
    \label{fig:heat}
\end{figure}

Figure \ref{fig:heat} further illustrates that models trained with LANE attend more strongly to the marked token, suggesting that adversarial negatives shift attention away from irrelevant contextual cues and toward the lexical element being disambiguated. This demonstrates that the benefits of LANE are not merely representational but also functional, influencing how the model allocates its focus during inference.

\begin{figure}
    \centering
    \includegraphics[width=\linewidth]{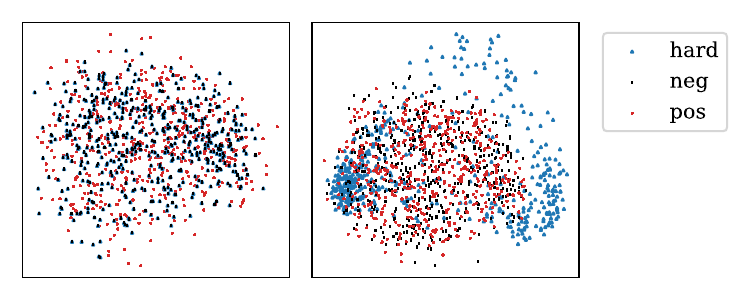}
    \caption{PCA for representations learned under a traditional regime (left) and under \ac{lane} (right). Adversarial negative examples are placed further apart in the embedding space.}
    \label{fig:pca}
\end{figure}

Finally, Figure \ref{fig:pca} shows that LANE yields more isotropic and semantically organized vector spaces. By evenly distributing representations and increasing separation among adversarial examples, the method promotes a more meaningful use of the embedding space.

In summary, LANE enhances both lexical sensitivity and representational structure in contextual models. Beyond improving WSD and cross-lingual generalization, it offers a principled mechanism for aligning neural optimization with linguistic distinctions—an essential step toward models that genuinely encode meaning rather than memorizing context.

\section{Conclusion}
\label{sec:conclusion}
We presented LANE, a simple and computationally efficient method for generating adversarial negatives in lexical representation tasks. By focusing on the target word rather than the surrounding context, LANE encourages models to encode fine-grained lexical distinctions, resulting in more robust and discriminative representations. Our experiments demonstrate that LANE consistently improves performance across monolingual and multilingual word sense disambiguation benchmarks, including out-of-distribution settings, without requiring complex hyperparameter tuning.

Importantly, we show that traditional approaches using marked words alone often fail to induce representations that prioritize lexical meaning, instead relying heavily on sentence-level contextual cues. LANE addresses this limitation by explicitly enforcing word-level separability, producing embeddings that better reflect true semantic distinctions.

Overall, LANE provides a simple, model-agnostic, and effective strategy for enhancing lexical semantic differentiation, improving generalization, and strengthening interpretability. Its ease of integration into existing frameworks makes it a practical tool for a wide range of lexical representation tasks, including cross-lingual and low-resource scenarios.

\section*{Limitations}
\label{sec:limitations}

A potential limitation of LANE lies in its word-substitution mechanism. If a randomly selected substitute is a synonym or near-synonym of the target word (e.g., replacing ``buy'' with ``purchase'' in the same sentence), the resulting pair may constitute a false negative, encouraging the model to separate semantically identical contexts. Although such occurrences are rare, the current formulation does not explicitly prevent them, representing an area for future refinement.
Additionally, LANE assumes simple tokenization by spaces, which may be insufficient for languages with complex word formation or rich morphology, potentially affecting its ability to accurately distinguish lexemes in such languages.


\bibliography{anthology,custom}

\clearpage
\section{Scheduling Adversarial Negatives}
\label{sec:schedule}
The early introduction of adversarial examples traps the model in a local optima, as we show in the Figure below.

\begin{figure}[htpb]
    \centering
    \includegraphics[width=\linewidth]{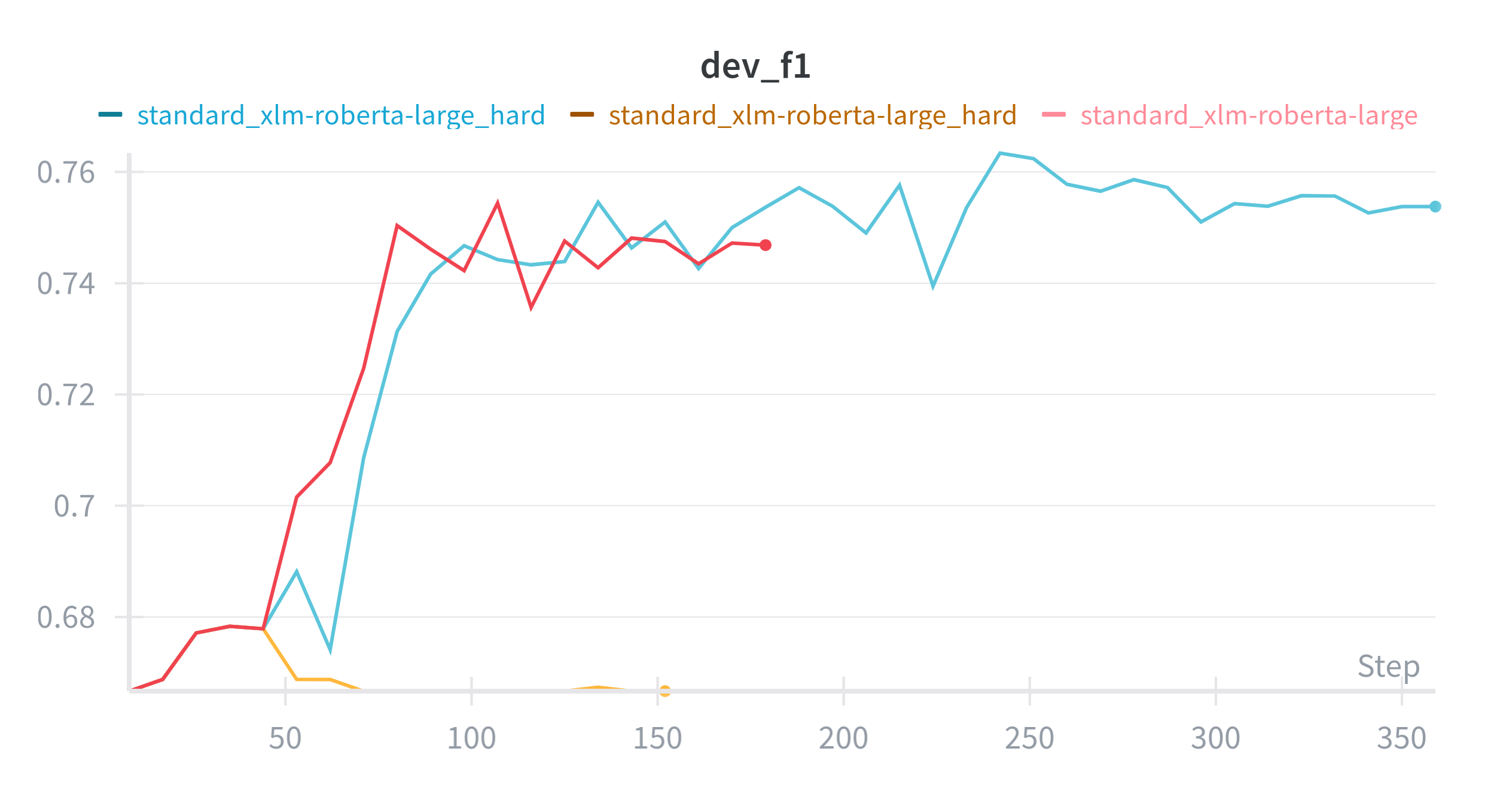}
    \caption{F1 score in the development data with different training settings.}
    \label{fig:placeholder}
\end{figure}

The red line represent a training without adversarial examples, the yellow line the training with an early insertion of adversarial examples (first epoch), and the blue line the scheduled insertion of adversarial examples.

\section{Datasets Licensing and Usage}
\label{app:datasets}
The datasets we compiled from existing resources (MASC, SEMCOR, FEWS), are under CC BY-SA 4.0. WIC and XL-WIC belongs to the original licensing CC BY-NC 4.0.

This data is intended to use for word in context differentiation, similar to the word-in-context tasks. In the table below we list the statistics of each dataset.
\begin{table}[h!]
\centering
\begin{tabular}{lrr}
\hline
\textbf{Dataset} & \textbf{Split} & \textbf{Instances} \\
\hline
SEMCOR & Train & 33,313 \\
SEMCOR & Dev & 7,000 \\
SEMCOR & Test & 9,674 \\\hline
MASC & Train & 7,280 \\
MASC & Dev & 7,000 \\
MASC & Test & 2,968 \\\hline
FEWS & Train & 132,237 \\
FEWS & Dev & 7,000 \\
FEWS & Test & 29,708 \\\hline
DWUG & Train & 32,424 \\
DWUG & Dev & 7,000 \\
DWUG & Test & 6,993 \\
\hline
\end{tabular}
\caption{Dataset split sizes for SemCor, MASC, FEWS, and DWUG.}
\label{tab:dataset_splits}
\end{table}

\section{Compute Costs}
To train the models on SEMCOR, MASC, DWUG, WIC, and FEWS we used a NVIDIA V100 32GB for less then 4 hours. The XL-WIC takes approximately 35 hours in the same GPU.

\section{Use of AI}
We use AI as a code assistant and as a writing assistant, for improving grammar and readability.

\end{document}